%% file: wipes.tex
\documentclass[10pt,twocolumn,letterpaper]{article}
\usepackage[pagenumbers]{iccv} 
\usepackage[accsupp]{axessibility}  


\definecolor{iccvblue}{rgb}{0.21,0.49,0.74}
\usepackage[pagebackref,breaklinks,colorlinks,allcolors=iccvblue]{hyperref}

\usepackage{array}
\usepackage{float}
\usepackage{xcolor}
\usepackage{colortbl}
\usepackage{multirow}
\usepackage{tabularx}
\usepackage{booktabs}
\usepackage{diagbox}
\usepackage{caption}
\usepackage{amsmath}
\usepackage{subcaption}
\newcolumntype{C}{>{\centering\arraybackslash}X}

\def\paperID{8425} 
\def\confName{ICCV}
\def\confYear{2025}

\title{WIPES: Wavelet-based Visual Primitives\thanks{The work was supported in part by NSFC Grant 62025108 and Tencent Rhino-Bird Joint Research Program RBFR2024009.}}

\author{
Wenhao Zhang$^{1,\dagger}$, Hao Zhu$^{1,\dagger}$, Delong Wu$^{1,\dagger}$, Di Kang$^2$, Linchao Bao$^2$, Xun Cao$^{1,*}$, Zhan Ma$^{1}$\\
$^1$ Nanjing University, $^2$ Tencent\\
$^\dagger$ Equal contribution. * Corresponding author: {\tt caoxun@nju.edu.cn}
}

\begin{document}
\maketitle

\begin{abstract}
Pursuing a continuous visual representation that offers flexible frequency modulation and fast rendering speed has recently garnered increasing attention in the fields of 3D vision and graphics. However, existing representations often rely on frequency guidance or complex neural network decoding, leading to spectrum loss or slow rendering. To address these limitations, we propose \textbf{WIPES}, a universal \textbf{W}avelet-based v\textbf{I}sual \textbf{P}rimitiv\textbf{ES} for representing multi-dimensional visual signals. Building on the spatial-frequency localization advantages of wavelets, WIPES effectively captures both the low-frequency ``forest" and the high-frequency ``trees." Additionally, we develop a wavelet-based differentiable rasterizer to achieve fast visual rendering. Experimental results on various visual tasks, including 2D image representation, 5D static and 6D dynamic novel view synthesis, demonstrate that WIPES, as a visual primitive, offers higher rendering quality and faster inference than INR-based methods, and outperforms Gaussian-based representations in rendering quality. Our project page: \url{https://mitnku.github.io/WIPES/}

\end{abstract}

\begin{figure}[!htbp]
    \centering
    \begin{subfigure}{\linewidth}
        \centering
        \includegraphics[width=0.92\textwidth]{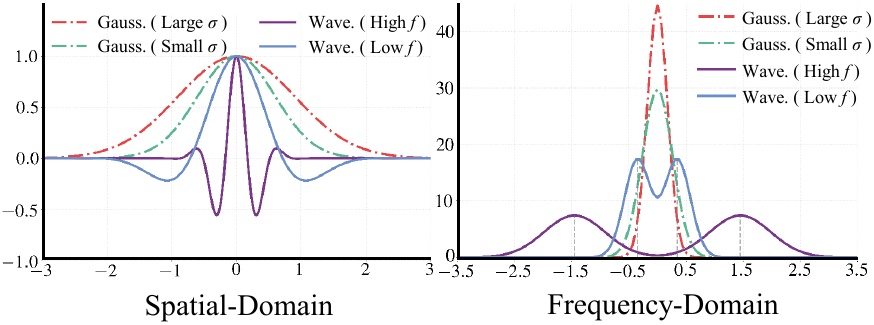}
        \caption{Comparison of Gaussians and WIPES} 
        \label{fig:time_freq_domain}
    \end{subfigure}
    
    \begin{subfigure}{\linewidth}
        \centering
        \includegraphics[width=0.92\textwidth]{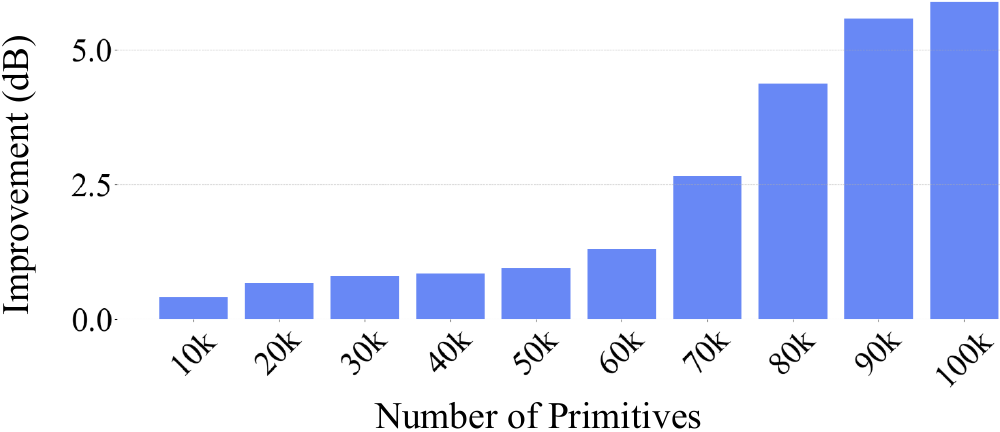}
        \caption{PSNR improvement of WIPES over Gaussian primitives \cite{Zhang2024GaussianImage}}
        \label{fig:point_number_compare}
    \end{subfigure}
    
    \begin{subfigure}{\linewidth}
        \centering
        \includegraphics[width=0.92\textwidth]{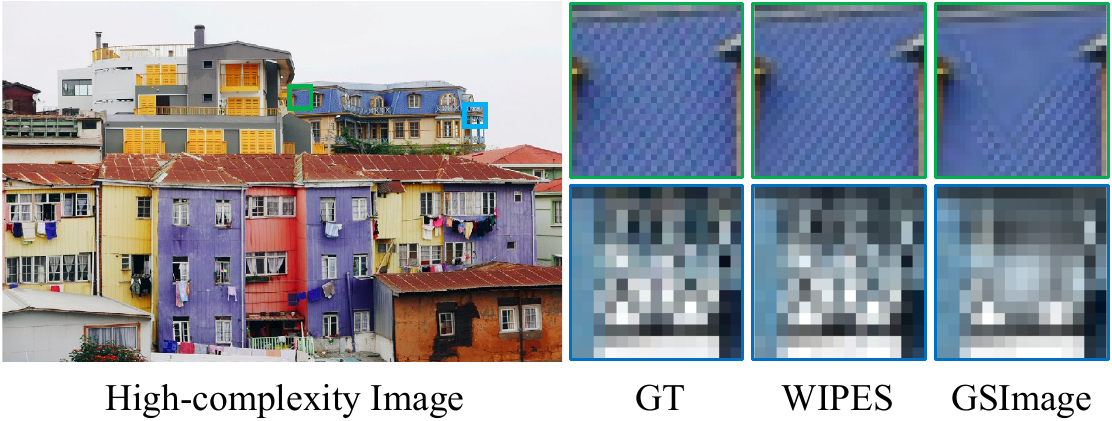}
        \caption{Performance comparison of WIPES and Gaussian primitives \cite{Zhang2024GaussianImage}}
        \label{fig:dataset_comparison}
    \end{subfigure}
    \caption{Spatial-Frequency Localization Benefits of WIPES. (a) shows the spatial and frequency distributions of Gaussian and WIPES primitives with various parameters. Adjusting $\sigma$ only affects the Gaussian frequency bandwidth, unable to overcome its low-frequency constraint. In contrast, WIPES flexibly modulates frequency, making it advantageous for complex visual tasks. (b) displays image fitting on the Kodak dataset, where WIPES achieves higher PSNR with the same primitives count, and this advantage grows with more primitives. (c) illustrates that Gaussian primitives smooth complex textures due to low-frequency limitation, while WIPES accurately reconstructs rooftop patterns and railings through adaptable frequency properties.}
    \vspace{-1.2em}
    \label{fig:gaussian_vs_wavelet}
\end{figure}

\section{Introduction}
\label{sec:Introduction}

Representing visual signals as continuous functions has garnered increasing attention in recent years. Owing to the inherent advantages of end-to-end differentiability and scalable querying, these continuous representations can be seamlessly integrated with various imaging and rendering functions. This integration facilitates the optimization of inverse problems, whether the resolution is down-sampled or the dimensionality is reduced. Since these representations are optimized independently for different scenarios, pursuing a universal visual primitive, which offers complex frequency compatibility and fast rendering speed, is of paramount importance.

However, existing representations suffer from spectral loss or slow rendering. For instance, the representational capacity of popular implicit neural representations (INRs) \cite{Sitzmann2020PeriodicINR} is often constrained by spectral bias \cite{Rahaman2019SpectralBias}, only the frequencies defined by hyperparameters are effectively learned \cite{Tancik2020FourierFeatures,yuce2022structured,zhu2024finer++}, resulting in the loss of information in other frequency bands. Furthermore, the complex network architectures of INRs require substantial computational resources \cite{Mildenhall2020NeRF} to decode signal attributes, thereby impeding processing speed across various visual tasks. Explicit representations, among which the 3D Gaussian Splatting (3DGS) \cite{Kerbl2023GaussianSplatting} specifically models the visual world by employing multiple Gaussian functions. However, the inherent low-pass nature of Gaussian functions introduces inaccuracies in capturing high-frequency details. To mitigate these low-pass characteristics, frequency regularization \cite{Zhang2024FreGS} or modulation losses \cite{Hamdi2024GES} are often employed, which necessitates extensive expertise to tune hyperparameters for different scenes.

\begin{figure*}[!htbp]
    \centering
    \includegraphics[width=1.0\textwidth]{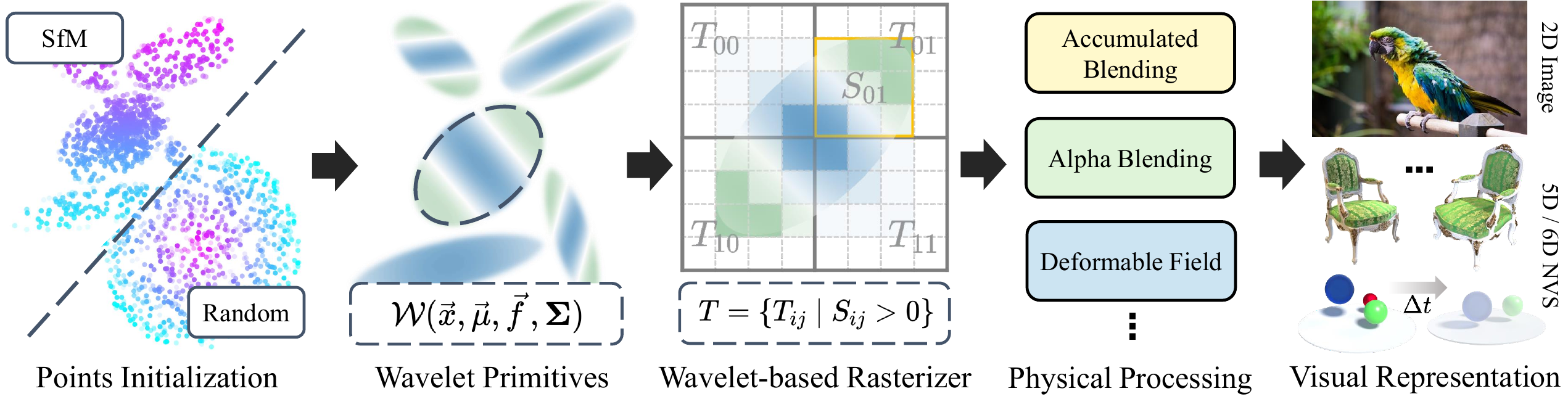}
    \caption{Pipeline of WIPES. WIPES can be seamlessly integrated into various visual reconstruction tasks, including 2D image representation, 5D static and 6D dynamic novel view synthesis (NVS).}
    \vspace{-0.6em}
    \label{fig:pipline}
\end{figure*}

These problems arise from a mismatch between the characteristics of the visual primitives used and the visual signals themselves. According to Marr's vision theory, visual signals are perceived hierarchically, and intensity changes are detected at multiple scales. Wavelet, designed for multi-scale signal analysis, aligns well with the human visual system and can capture both the ``forest" and the ``trees" in visual signals \cite{Graps1995Wavelets}. Consequently, we developed WIPES, a universal primitive for representing visual signals.

WIPES leverages the Morlet wavelet—known for its continuous differentiability and robust spatial-frequency localization—to effectively capture signals exhibiting significant frequency variations. We extend the Morlet wavelet by developing a wavelet primitives adaptable to arbitrary dimensions, making it well-suited for various visual signals in real-world applications (\eg, 2D images, 5D static and 6D dynamic radiance fields). Additionally, we develop a fast, differentiable rasterizer for WIPES by utilizing an efficient GPU sorting algorithm \cite{Lassner2021Pulsar}. To mitigate the substantial computational overhead associated with automatic differentiation during training, we derive explicit gradients for all parameters in both forward and backward propagations. Our rasterizer effectively replaces Gaussian-based rasterizers, enabling seamless integration with different loss functions \cite{Zhang2024FreGS, Chung2024DepthRegularized3DGS} and subsequent visual tasks \cite{Zhang2024GaussianImage, Yang2024Deformable3DGaussians, Zha2024R2Gaussian} built upon Gaussian splatting techniques.


In summary, we make the following contributions:
\begin{itemize}
    \item We design a universal differentiable wavelet primitive that adaptively models various frequency features in real-world scenes without relying on handcrafted frequency guidance.
    \item We develop a fast differentiable rasterizer for our WIPES, enabling anisotropic splatting and efficient backpropagation across different dimensions to achieve general and effective visual representation.
    \item The proposed wavelet-based approach benefits superior reconstruction performance across multiple task dimensions compared to other visual representations.
\end{itemize}


\section{Related Work}
\label{sec:related}

\noindent\textbf{Implicit Neural Representations (INRs)} play a pivotal role in modeling continuous functions of signals found in images, scenes and videos \cite{Lindell2021BACON}. They leverage MLPs to accurately map spatial coordinates to their corresponding values \cite{Mildenhall2020NeRF, Sitzmann2020PeriodicINR}. This capability enables INRs to be applied in various domains, including novel view synthesis \cite{Mildenhall2020NeRF}, free-hand 3D/4D echocardiography \cite{shen2024cardiacfield, shen2024continuous}, microscopy imaging \cite{Zhu2022DNF}, and nano-optics design \cite{chen2020inverse}.

However, INRs are constrained by spectral bias, favoring low-frequency information over high-frequency details \cite{Rahaman2019SpectralBias}. To mitigate this bias, strategies like positional encoding \cite{Tancik2020FourierFeatures} and specifically activation functions \cite{Sitzmann2020PeriodicINR, ramasinghe2022beyond, Saragadam2023WIRE, liu2024finer} have been developed. Despite these advancements, the spectral bias remains fundamentally unaddressed and is merely transformed into a ``frequency''-specified spectral bias \cite{zhu2024finer++}. Additionally, INRs suffer from slow training and inference due to their reliance on computationally intensive MLPs. To enhance efficiency, recent approaches have integrated explicit storage with implicit decoding, including sparse voxel grids \cite{Yu2021Plenoxels}, hash tables \cite{Muller2022InstantNeuralGraphics, zhu2024disorder, zhu2025rhino}, and multi-scale/grid INRs \cite{Saragadam2022MINER, reiser2021kilonerf,zhu2023pyramid}. While these approaches offer improvements, they introduce additional complexity, making it challenging to balance memory consumption with real-time performance. Furthermore, INRs are highly sensitive to parameter changes. Modifications to the input or internal parameters can lead to significant variations in the output, complicating the process of making local adjustments \cite{Martel2021Acorn}. Although strategies such as explicit encoding and latent disentanglement \cite{Kohli2020SemanticINR, Wu2025NeuralPoissonSolver} have been proposed to address these issues, achieving precise control over local features in INR-based approaches remains challenging.

\vspace{0.1cm}
\noindent\textbf{Explicit Signal Representations with Gaussian Primitives}, conversely, overcome several INR limitations and are growing in popularity in computer vision and graphics. They use multiple independent continuous Gaussian functions to render scenes, which offers marked benefits like enhanced expressiveness and editability, and superior fitting efficiency than conventional INRs.

These Gaussian-based approaches have achieved top-tier performance in rendering quality and speed \cite{Kerbl2023GaussianSplatting}. Early research on splatting introduced Gaussian primitives for scene representation \cite{Zwicker2001SurfaceSplatting, Zwicker2001EWA}. Recent advancements such as FreGS \cite{Zhang2024FreGS}, mip-splatting \cite{Yu2024MipSplatting}, and scaffold-GS \cite{Lu2024ScaffoldGS} enhance frequency management, mitigate aliasing, and incorporate voxel-based initialization, thereby demonstrating the robustness and efficiency of Gaussian-based approaches. 

However, Gaussian primitives are constrained by their inherent low-pass filtering characteristics, which are effective for smooth signals but hinder the capture of high-frequency details like sharp edges or intricate textures. Real-world scenes frequently contain abrupt changes, such as the complex textures of grass or nylon fabric. Accurately representing these details requires multiple small-scale Gaussians \cite{Huang20242DGaussianSplatting, Hamdi2024GES}, which significantly increase model complexity, thereby reducing scalability and efficiency.

\vspace{0.1cm}
\noindent\textbf{Wavelet-based Signal Representations} have been proven to be highly effective in analyzing complex signals within vision tasks. Traditional Fourier approaches use sine functions for their smoothness and clear characteristics in the Fourier domain. However, they decompose signals into global sine bases, which lack localization in both space and frequency \cite{Mallat1999WaveletTour}. This limitation makes Fourier approaches less suitable for visual data with local features and abrupt changes. In contrast, wavelets provide optimal localization in both domains, effectively capturing local high-frequency variations and sudden signal changes while maintaining a compact representation. Recent studies have leveraged wavelets to enhance vision models. MFN \cite{Fathony2021MFN} employs wavelets as nonlinear filters to approximate complex functions with fewer parameters, improving frequency information representation. TriNeRFLet \cite{Khatib2024TriNeRFlet} integrates wavelets into Triplane’s multi-scale representations, enhancing cross-scale information sharing and high-frequency regularization for better 3D reconstructions. WavePlanes \cite{Azzarelli2023WavePlanes} uses wavelet compression on the spatiotemporal feature planes of dynamic NeRFs, reducing model size and enhancing high-frequency details.

However, recent wavelet-inspired representations are implicitly integrated into neural networks, which significantly reduce the rendering speed. On the contrary, the proposed wavelet-based primitives operate on the visual signal explicitly and directly, thus improving the efficiency while maintaining the advantages of spatial-frequency location.


\section{Preliminaries}
\label{sec:methods}
\subsection{3D Gaussian Splatting}
\label{sec::3d with gs}

3DGS \cite{Kerbl2023GaussianSplatting} explicitly represents 3D scenes using Gaussian primitives, enabling efficient scene reconstruction. Specifically, each Gaussian primitive is parameterized by a spatial mean vector $\vec{\mu}$, a covariance matrix $\mathbf{\Sigma}$ (expressed as $\mathbf{\Sigma} = \mathbf{R}\mathbf{S}\mathbf{S}^{\top}\mathbf{R}^{\top}$, where $\mathbf{R}$ denotes rotation and $\mathbf{S}$ denotes scaling), an opacity value $\alpha$, and color attributes encoded by spherical harmonic coefficients $c$. In the world space, the Gaussian primitive centered at $\vec{\mu}$ with $\mathbf{\Sigma}$ is defined as:
\begin{equation}
    \mathcal{G}(\vec{x},\vec{\mu},\mathbf{\Sigma}) = e^{-\frac{1}{2} (\vec{x} - \vec{\mu})^\top \mathbf{\Sigma}^{-1} (\vec{x} - \vec{\mu})},
    \label{eq:formulation of Gaussian}
\end{equation}
where $\vec{x} $ and $\vec{\mu}$ are column vectors in the world space, denoted as $ (x,y,z)^{\top}$ and $(\mu_x, \mu_y, \mu_z)^{\top}$, respectively, the covariance matrix $\mathbf{\Sigma}$ is a 3$\times$3 symmetric matrix. 




When employing the splatting method for rendering, the 3D reconstruction kernel must be integrated along the $z$-axis to evaluate its contribution to each pixel on the image plane. Since integrating a 3D Gaussian function along a given axis still results in a Gaussian function, we can effectively transform a 3D Gaussian $\mathcal{G}(\vec{x})$ into a 2D Gaussian $\mathcal{G}'(\vec{x}')$ on the image plane using the ray-coordinate transformation described in \cite{Zwicker2001EWA}:

\begin{equation}
\int_{\mathbb{R}}\mathcal{G}(\vec{x},\vec{\mu},\mathbf{\Sigma})dz=\mathcal{G}'(\vec{x}',\vec{\mu}',{{\mathbf{\Sigma}'}}),
\label{eq:: integration of 3DGS}
\end{equation}
where $\vec{x}'=(x, y)^{\top}$ and $\vec{\mu}' = (\mu_x, \mu_y)^{\top}$ represent coordinates in the projected 2D image plane. To achieve accurate perspective projection, the original covariance matrix $\mathbf{\Sigma}$ undergoes transformation through the world-to-camera transformation matrix $\mathbf{W}$, followed by projection via the local affine Jacobian matrix $\mathbf{J}$. This yields a ray-space covariance matrix $\hat{\mathbf{\Sigma}}=\mathbf{J}\mathbf{W}\mathbf{\Sigma}\mathbf{W}^{\top}\mathbf{J}^{\top}$. Ultimately, the 2D covariance matrix $\Sigma'$ in image space can be easily obtained by taking the upper left 2$\times$2 submatrix of $\hat{\mathbf{\Sigma}}$.

Finally, the pixel values on the 2D image are obtained by $\alpha$-blending:
\begin{equation}
C({\vec{x}'})=\sum_{i\in{N}}c_i\alpha_i\mathcal{G}_{i}'(\vec{x}',\vec{\mu}',\mathbf{\Sigma}')\prod_{j=1}^{i-1}(1-\alpha_j\mathcal{G}_{j}'(\vec{x}',\vec{\mu}',\mathbf{\Sigma}')),
\label{eq::a-blending gs}
\end{equation}
where $\vec{x}'$ denotes the queried pixel position and $N$ represents the set of sorted 2D Gaussians associated with $\vec{x}'$.


\begin{table*}[!t]
    \centering
    \setlength\tabcolsep{12pt}
    \caption{Quantitative comparison of image fitting on Kodak and DIV2K datasets. We color code each cell as \colorbox{pink!95}{best} and \colorbox{pink!30}{second best}.}
    \vspace{-0.8em}
    \label{tab:2d_exp}
    \small
    \begin{tabularx}{\textwidth}{@{}p{2.5cm}|*{4}{c}|*{4}{c}@{}} 
    \toprule
    \multirow{2}{*}{Methods $\mid$ Metrics} & \multicolumn{4}{c|}{\textbf{Kodak}} & \multicolumn{4}{c}{\textbf{DIV2K}} \\ 
     & PSNR$\uparrow$ & SSIM$\uparrow$ & LPIPS$\downarrow$ & FPS$\uparrow$ & PSNR$\uparrow$ & SSIM$\uparrow$ & LPIPS$\downarrow$ & FPS$\uparrow$ \\ 
    \midrule
    WIRE \cite{Saragadam2023WIRE} & 38.02 & 0.9541 & 0.0477 & 19.55 & 33.20 & 0.9565 & 0.1303 & 11.63 \\
    GSImage-RS \cite{Zhang2024GaussianImage} & 43.09 & 0.9983 & 0.0216 & \cellcolor{pink!95}2129.18 & 39.85 & \cellcolor{pink!95}0.9978 & 0.0371 & \cellcolor{pink!30}2190.40 \\
    GSImage-Cholesky & 44.06 & 0.9985 & 0.0188 & \cellcolor{pink!30}2061.99 & 39.53 & 0.9975 & 0.0430 & \cellcolor{pink!95}2248.74 \\
    \textbf{WIPES-RS} &\cellcolor{pink!30} 45.76 & \cellcolor{pink!95}0.9988 & \cellcolor{pink!30}0.0132 & 1708.28 & \cellcolor{pink!95}40.34 & \cellcolor{pink!30}0.9976 & \cellcolor{pink!95}0.0360 & 1817.48 \\
    \textbf{WIPES-Cholesky} & \cellcolor{pink!95}45.87 & \cellcolor{pink!30}0.9987 & \cellcolor{pink!95}0.0120 & 1778.75 & \cellcolor{pink!30}40.32 &\cellcolor{pink!95} 0.9978 & \cellcolor{pink!30} 0.0361 & 1830.10 \\
    \bottomrule
    \end{tabularx}
\end{table*}


\begin{figure*}[!t]
    \centering
    \vspace{-0.6em}
    \includegraphics[width=1.0\textwidth]{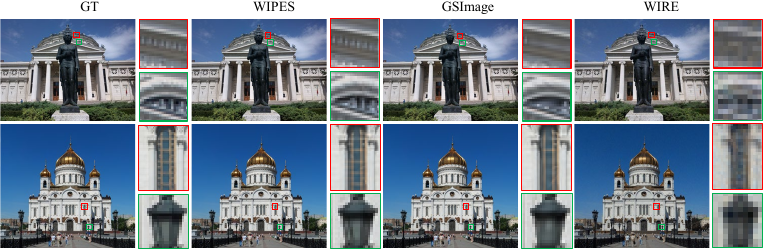}
    \vspace{-1.8em}
    \caption{Qualitative comparison on Kodak and DIV2K datasets. Our approach effectively captures both low- and high-frequency details.}
    \vspace{-0.6em}
    \label{fig:2d_exp}
\end{figure*}

\subsection{Expansion of 3D Gaussian Splatting}
\label{sec::Expansion of 3dgs}

Gaussian-based representations have demonstrated exceptional performance not only in 3D scene signal modeling but also in various other domains. For instance, Gaussian Image \cite{Zhang2024GaussianImage} explicitly parameterizes 2D image signals using 2D Gaussian primitives defined directly on the image plane, replacing traditional $\alpha$-blending with a simplified weighted summation mechanism:

\begin{equation}
C({\vec{x}'})=\sum_{i\in{N}}c_i\alpha_i\mathcal{G}_i'(\vec{x}', \vec{\mu}', \mathbf{\Sigma}').
\label{sum of gs}
\end{equation}

Moreover, to handle dynamic 3D content, D-3DGS \cite{Yang2024Deformable3DGaussians} employs an MLP network $\mathcal{F}_\theta$ to model the deformation field of 3D Gaussian primitives. Thus, the primitives efficiently adapt to dynamic variations over time, yielding a high-fidelity representation of temporally-varying 3D scenes:

\begin{equation}(\delta{\vec{x}},\delta\mathbf{\Sigma})=\mathcal{F}_\theta(\gamma(\vec{x}),\gamma(t)),
\end{equation}
where $\gamma$ denotes positional encoding \cite{Mildenhall2020NeRF}.

\section{Wavelet-based Visual Primitives}
\label{sec::diff wavelet kernel}
\subsection{3D Splatting with Wavelet Primitives}

The wavelet transform decomposes a signal into a linear combination of translated and scaled short oscillatory pulses, effectively capturing localized variations in both spatial and frequency domains. Due to their compact support in these domains, wavelets achieve superior accuracy in signal approximation \cite{Devore1998Nonlinear}. This property substantiates the feasibility of using differentiable wavelet basis functions in our methodology. Inspired by the Morlet wavelet, we obtain the 3D wavelet primitive by applying cosine modulation to the 3D Gaussian primitive. A 3D wavelet primitive is described by: 
\begin{equation}
    \mathcal{W}(\vec{x},\vec{\mu},\vec{f}, \mathbf{\Sigma}) = \frac{1}{2}[\cos(\vec{f} \cdot (\vec{x}-\vec{\mu})) + 1] \, \mathcal{G}(\vec{x},\vec{\mu},\mathbf{\Sigma}),
    \label{eq:formulation of wavelets}
\end{equation}
where $\vec{f}=(f_x, f_y, f_z)^{\top}$ is the frequency of cosine modulation. Notably, the Gaussian primitive can be considered a subset of the proposed 3D wavelet primitive by setting the parameter $\vec{f}$ to $\vec{0}$.

Integrating the 3D wavelet primitive along the $z$-axis produces results similar to those described in \cref{eq:: integration of 3DGS}:
\begin{equation}
\int_{\mathbb{R}}\mathcal{W}(\vec{x},\vec{\mu},\vec{f}, \mathbf{\Sigma})dz=\mathcal{W}'(\vec{x}',\vec{\mu}',\vec{f}', \mathbf{\Sigma}'),
    \label{eq:integartion of wavelets}
\end{equation}
where $\vec{f}'=(f_x, f_y)^{\top}$ and $\vec{x}'-\vec{\mu}'$ is the same as the 3D Gaussian primitive in \cref{eq:: integration of 3DGS}.

\begin{figure}[!htbp]
    \centering
    \includegraphics[width=\linewidth]{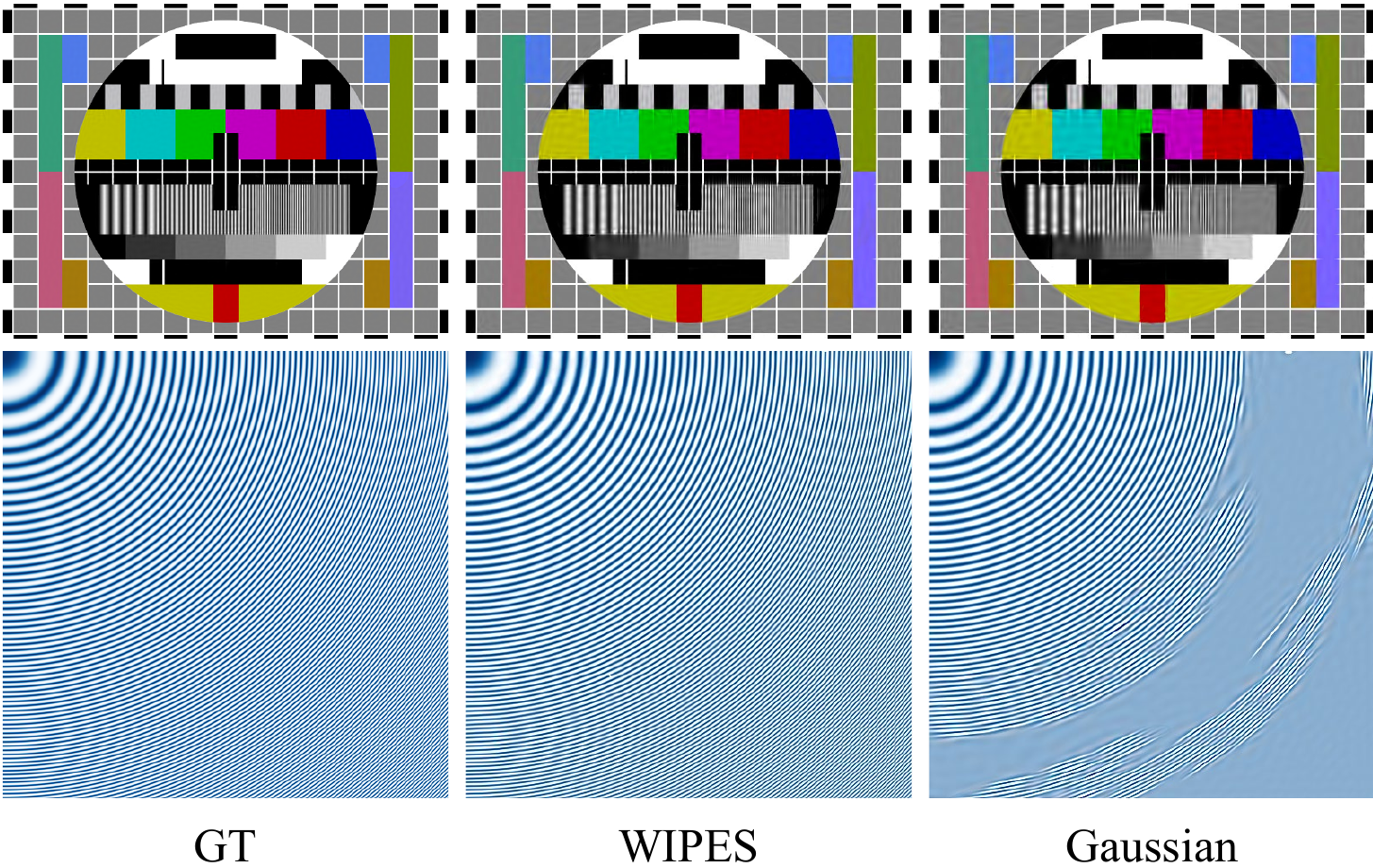}
    \caption{Comparisons of Gaussian and wavelet primitives for representing complex texture patterns. Results indicate that wavelet primitives effectively overcome the low-frequency bias of Gaussian primitives and demonstrate excellent performance in high-frequency texture regions.}
    \vspace{-1.2em}
    \label{fig:toy_exp}
\end{figure}


The projected 2D covariance matrix $\mathbf{\Sigma}’$ for each wavelet primitive is computed via a similar spatial transformation used in projecting Gaussian primitives from 3D world space to the 2D image plane. Specifically, the covariance matrix $\mathbf{\Sigma}$ and modulation frequency vector $\vec{f}$ of the wavelet primitive undergo the identical ray-space transformation defined by matrices $\mathbf{W}$ and $\mathbf{J}$, resulting in the transformed frequency vector $\vec{f}’$ and the projected covariance matrix $\mathbf{\Sigma}’$. 


For rasterization, we follow a similar pipeline as in 3D GS while substituting the primitives with our proposed 3D wavelets. In the end, the volumetric $\alpha$-blending can be easily obtained by replacing the Gaussian primitive $\mathcal{G}'$ in \cref{eq::a-blending gs} with the wavelet primitive $\mathcal{W}'$.
\begin{table*}[!t]
    \centering
    \setlength\tabcolsep{4.85pt}
    \caption{Quantitative comparison of static novel view synthesis on Mip-NeRF360, Tanks $\text{\&}$ Temples and DeepBlending datasets. Results marked with $\dagger$ were adopted from the 3DGS paper \cite{Kerbl2023GaussianSplatting}, all others come from our experiments.}
    \vspace{-0.6em}
    \label{tab:5d_exp}
    \small
    \begin{tabularx}{\textwidth}{@{}p{2.5cm}|*{4}{c}|*{4}{c}|*{4}{c}@{}} 
    \toprule
    \multirow{2}{*}{Methods $\mid$ Metrics} & \multicolumn{4}{c|}{\textbf{Mip-NeRF360}} & \multicolumn{4}{c|}{\textbf{Tanks \& Temples}} & \multicolumn{4}{c}{\textbf{DeepBlending}} \\ 
     & PSNR$\uparrow$ & SSIM$\uparrow$ & LPIPS$\downarrow$ & FPS$\uparrow$ & PSNR$\uparrow$ & SSIM$\uparrow$ & LPIPS$\downarrow$ & FPS$\uparrow$ & PSNR$\uparrow$ & SSIM$\uparrow$ & LPIPS$\downarrow$ & FPS$\uparrow$ \\ 
    \midrule
    Plenoxels \cite{Yu2021Plenoxels} & 23.32 & 0.646 & 0.451 & 6.79 & 21.08 & 0.719 & 0.379 & 13 & 23.06 & 0.795 & 0.510 & 11.2 \\
    INGP \cite{Muller2022InstantNeuralGraphics} & 25.19 & 0.667 & 0.440 & 11.7 & 21.72 & 0.723 & 0.330 & 17.1 & 23.62 & 0.797 & 0.423 & 3.26 \\
    Mip-NeRF360$^\dagger$ \cite{Barron2022MipNeRF360} & \cellcolor{pink!95}27.69 & 0.792 & 0.237 & 0.06 & 22.22 & 0.759 & 0.257 & 0.14 & 29.40 & 0.951 & 0.245 & 0.09 \\
    3DGS \cite{Kerbl2023GaussianSplatting} & 27.28 & \cellcolor{pink!30}0.811 & \cellcolor{pink!30}0.220 & \cellcolor{pink!95}118.2 & \cellcolor{pink!30}23.35 & \cellcolor{pink!30}0.835 & \cellcolor{pink!30}0.182 & \cellcolor{pink!95}162.4 & \cellcolor{pink!30}29.43 & \cellcolor{pink!30}0.898 & \cellcolor{pink!30}0.246 & \cellcolor{pink!95}124.1 \\
    \textbf{WIPES} & \cellcolor{pink!30}27.55 & \cellcolor{pink!95}0.816 & \cellcolor{pink!95}0.215 & \cellcolor{pink!30}95.7 & \cellcolor{pink!95}23.78 & \cellcolor{pink!95}0.852 & \cellcolor{pink!95}0.171 & \cellcolor{pink!30}126.3 & \cellcolor{pink!95}29.82 & \cellcolor{pink!95}0.907 & \cellcolor{pink!95}0.238 & \cellcolor{pink!30}92.6 \\
    \bottomrule
    \end{tabularx}
\end{table*}

  \begin{figure*}[!t]
    \centering
    \vspace{-1em}
    \includegraphics[width=\textwidth]{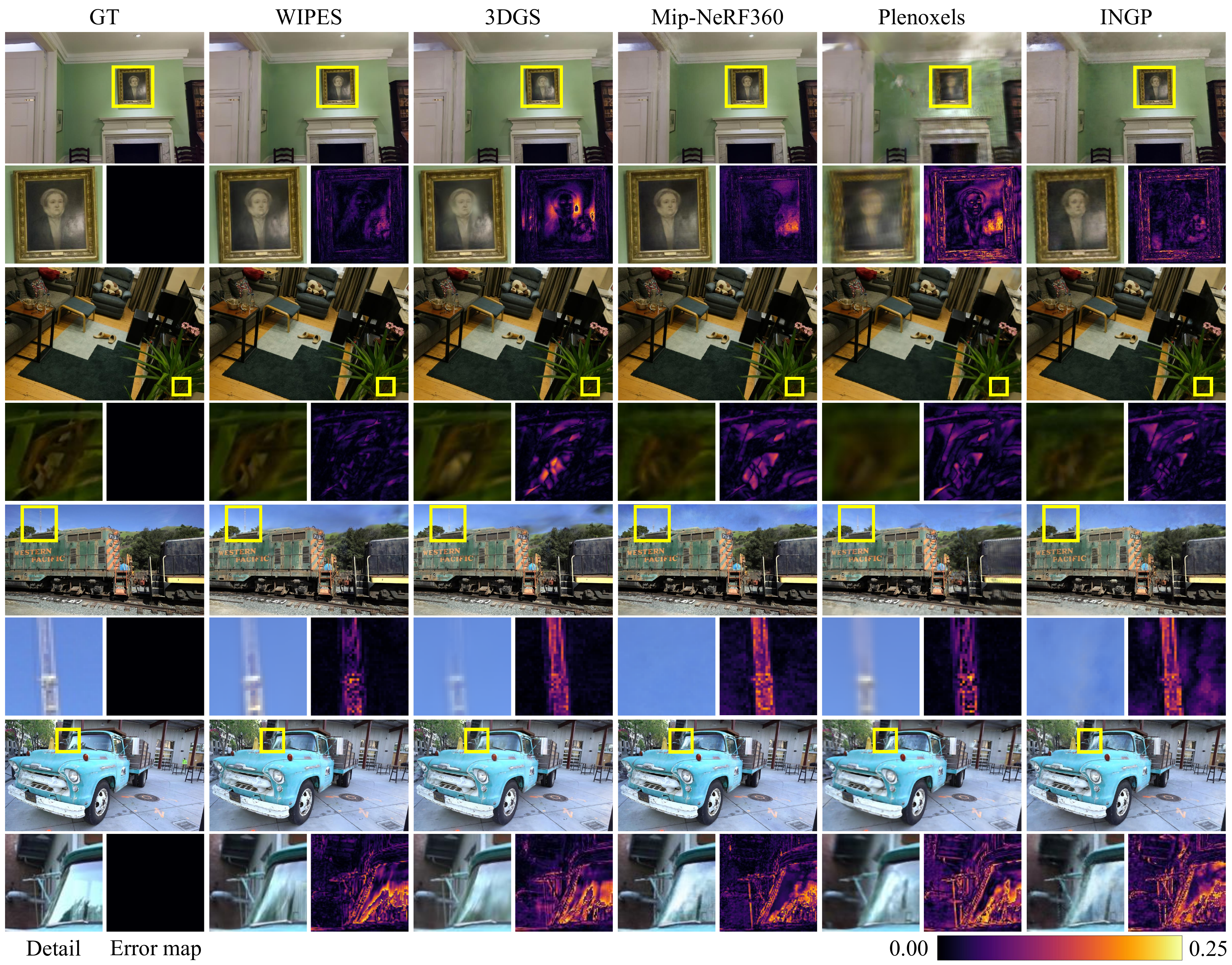}
    \vspace{-1.7em}
    \caption{Qualitative comparison of static novel view synthesis on Mip-NeRF360, Tanks $\text{\&}$ Temples and DeepBlending datasets. 
    }
    \vspace{-1.5em}
    \label{fig:5d_exp}
\end{figure*}

\subsection{Wavelet-Based Frequency Flexibility}
\label{sec::disussion of wavelet}

Gaussian functions inherently act as low-pass filters, as illustrated in \cref{fig:time_freq_domain}. Specifically, in the frequency domain, Gaussians with large variance exhibit narrow bandwidths concentrated around zero frequency, while those with small variance produce wider bandwidths. This bell-shaped Gaussian spectrum rapidly attenuates high-frequency signals, thus limiting its capacity to represent signals rich in high-frequency content. Consequently, a large number of Gaussian basis functions with smaller variances are inevitably required to accurately capture intricate high-frequency details.


In contrast, wavelets provide significantly enhanced frequency flexibility. By employing adaptive frequency modulation, wavelet primitives simultaneously encode both low- and high-frequency components within a single basis function. This modulation introduces a controllable shift in the wavelet spectrum, positioning the Gaussian-like envelope around a desired center frequency. The modulation frequency defines this spectral center, and the Gaussian envelope controls the bandwidth, ensuring optimal coverage of the spatial-frequency domain.


We validate this advantage by comparing Gaussian and wavelet primitives on two images exhibiting complex moiré patterns (\cref{fig:toy_exp}). Under identical conditions and equal numbers of primitives, Gaussian primitives consistently failed to represent detailed high-frequency structures, resulting in noticeable modeling voids. In contrast, the wavelet-based representation accurately captured intricate frequency transitions, effectively modeling sharp discontinuities and fine details. This result demonstrates that wavelets, through their inherent frequency adaptability, offer superior efficiency and accuracy in representing signals characterized by rapid spatial variations.

\subsection{Expansion of 3D Wavelet Splatting}


We further leverage wavelet primitives to enhance representation quality, especially for signals exhibiting complex spatial-frequency content. Specifically, we validate the effectiveness of wavelet primitives in two key applications: image fitting and dynamic novel view synthesis.

For the image fitting task, we replace the original 2D Gaussian primitives $\mathcal{G}'$ defined in \cref{sum of gs} with frequency-adaptive wavelet primitives $\mathcal{W}'$. This substitution introduces modulation frequencies $\vec{f}’$, significantly expanding frequency coverage compared to Gaussian-based models. In dynamic novel view synthesis, we further generalize wavelet primitives by making them temporally adaptive. The modified formulation is given by:
\begin{equation}(\delta{\vec{x}},\delta\vec{f}, \delta\mathbf{\Sigma})=\mathcal{F}_\theta(\gamma(\vec{x}),\gamma(t)).
\end{equation}

Comprehensive experimental validation demonstrating the superiority of our wavelet-based approach is provided in \cref{sec:exp}, confirming their ability to efficiently represent high-frequency details and complex temporal variations.

\section{Experiments}
\label{sec:exp}


Our experiments demonstrate the robustness and versatility of WIPES in representing real-world scenes, effectively capturing both high-frequency details (\eg, edge variations and intricate textures) and low-frequency contours. To validate WIPES as a generalized signal representation framework, we systematically evaluate its performance across three distinct tasks: 2D image fitting, 5D static novel view synthesis, and 6D dynamic novel view synthesis.

All experiments are consistently executed on a single NVIDIA A6000 GPU. To ensure a fair comparison, we adopt the same initialization schemes, optimization strategies, and hyperparameter configurations as those of the baseline approaches. Specifically, frequency modulation coefficients are randomly initialized in alignment with competing methods to exclude initialization bias. Additionally, our implementation builds upon the recent \texttt{gsplat} library \cite{Ye2024gsplat}, augmented by custom CUDA kernels that enable wavelet-based differentiable rasterization blending. This adaptation is critical for efficiently leveraging wavelet primitives in rendering tasks.

\subsection{2D Image Fitting}

\noindent\textbf{Configurations.} In this section, we utilized the Kodak \cite{HessKodakDataset} and DIV2K \cite{Agustsson2017NTIRE} datasets. The DIV2K dataset was downscaled by a factor of $2$ using bicubic interpolation, resulting in images with resolutions ranging from $408 \times 1020$ to $1020 \times 1020$. For our WIPES, we conducted experiments using the two covariance construction methods proposed in GSImage, namely Cholesky and RS. The frequency coefficients were randomly initialized with a normal distribution, while all other learnable parameters followed the initialization strategy described in GSImage \cite{Zhang2024GaussianImage}. To ensure a fair comparison with WIRE \cite{Saragadam2023WIRE} and GSImage \cite{Zhang2024GaussianImage}, all hyperparameters were kept consistent with those used in GSImage.

\noindent\textbf{Results.} \cref{tab:2d_exp} presents a quantitative comparison between our proposed WIPES and existing advanced INR-based and Gaussian-based approaches. Our approach achieves the best performance across all metrics, including PSNR, SSIM, and LPIPS. Additionally, as illustrated in \cref{fig:2d_exp}, WIPES effectively captures both low-frequency features (\eg, walls and gates) and high-frequency details (\eg, tiles and streetlamps) within the scenes. Although GSImage demonstrates the highest rendering speed, our approach maintains a competitive frame rate, ensuring efficient performance without sacrificing quality. This balance between high-quality image reconstruction and rendering efficiency underscores the practical effectiveness of our approach.


\begin{table}[!tbp]
    \centering
    \setlength\tabcolsep{8.6pt}
    \caption{Quantitative comparison on static novel view synthesis with GES \cite{Hamdi2024GES}, which incorporates frequency guidance, across the Mip-NeRF360, Tanks $\text{\&}$ Temples and DeepBlending datasets. GES(3e-4) uses the default parameters from the original GES paper, while GES(2e-4) adopts the same settings as 3DGS \cite{Kerbl2023GaussianSplatting}.}
    \label{tab:5d_vs_ges}
    \footnotesize
    \begin{subtable}{\linewidth}
        \centering
        \caption{Mip-NeRF360}
        \label{tab:mipnerf360}
        \begin{tabularx}{\linewidth}{@{}l|cccc@{}} 
            \toprule
            Methods $\mid$ Metrics & PSNR$\uparrow$ & SSIM$\uparrow$ & LPIPS$\downarrow$ & Points$\downarrow$ \\ 
            \midrule
            GES(3e-4) & 26.98 & 0.7208 & 0.3257 & \cellcolor{pink!30}1.521M \\
            \textbf{WIPES(3e-4)} & 27.11 & 0.7958 & 0.2518 & \cellcolor{pink!95}1.401M \\
            GES(2e-4) & \cellcolor{pink!30}27.36 & \cellcolor{pink!30}0.8148 & \cellcolor{pink!30}0.2172 & 3.055M \\
            \textbf{WIPES(2e-4)} & \cellcolor{pink!95}27.55 & \cellcolor{pink!95}0.8164 & \cellcolor{pink!95}0.2149 & 2.787M \\
            \bottomrule
        \end{tabularx}
    \end{subtable}
    
    \vspace{0.5em}
    
    \begin{subtable}{\linewidth}
        \centering
        \caption{Tanks $\text{\&}$ Temples}
        \label{tab:tanks_and_temples}
        \begin{tabularx}{\linewidth}{@{}l|cccc@{}} 
            \toprule
            Methods $\mid$ Metrics & PSNR$\uparrow$ & SSIM$\uparrow$ & LPIPS$\downarrow$ & Points$\downarrow$ \\ 
            \midrule
            GES(3e-4) & 23.44 & 0.8394 & 0.1954 & \cellcolor{pink!95}0.913M \\
            \textbf{WIPES(3e-4)} & \cellcolor{pink!30}23.76 & 0.8433 & 0.1914 & \cellcolor{pink!30}0.931M \\
            GES(2e-4) & 23.59 & \cellcolor{pink!30}0.8481 & \cellcolor{pink!30}0.1772 & 1.601M \\
            \textbf{WIPES(2e-4)} & \cellcolor{pink!95}23.78 & \cellcolor{pink!95}0.8523 & \cellcolor{pink!95}0.1709 & 1.591M \\
            \bottomrule
        \end{tabularx}
    \end{subtable}
    
    \vspace{0.5em}
    
    \begin{subtable}{\linewidth}
        \centering
        \caption{DeepBlending}
        \label{tab:deepblending}
        \begin{tabularx}{\linewidth}{@{}l|cccc@{}} 
            \toprule
            Methods $\mid$ Metrics & PSNR$\uparrow$ & SSIM$\uparrow$ & LPIPS$\downarrow$ & Points$\downarrow$ \\ 
            \midrule
            GES(3e-4) & 29.59 & 0.9053 & 0.2492 & \cellcolor{pink!30}1.243M \\
            \textbf{WIPES(3e-4)} & \cellcolor{pink!30}29.72 & \cellcolor{pink!30}0.9062 & 0.2501 & \cellcolor{pink!95}1.137M \\
            GES(2e-4) & 29.55 & 0.9041 & \cellcolor{pink!30}0.2434 & 2.059M \\
            \textbf{WIPES(2e-4)} & \cellcolor{pink!95}29.82 & \cellcolor{pink!95}0.9069 & \cellcolor{pink!95}0.2383 & 1.933M \\
            \bottomrule
        \end{tabularx}
    \end{subtable}
\end{table}

\subsection{5D Static Novel View Synthesis}
\label{sec:5d_exp}

\noindent\textbf{Configurations.} In this section, we evaluated WIPES on 13 scenes sourced from the Mip-NeRF360 \cite{Barron2022MipNeRF360}, Tanks $\text{\&}$ Temples \cite{Knapitsch2017TanksTemples}, and DeepBlending \cite{Hedman2018DeepBlending} datasets. Specifically, we set the learning rate for the frequency coefficients to $2.5 \times 10^{-3}$ in WIPES. Differentiable rasterization was implemented via custom CUDA kernels, while all other learnable parameters and experimental settings remained aligned with those of 3DGS \cite{Kerbl2023GaussianSplatting}. This configuration facilitated a direct comparison with state-of-the-art approaches including Plenoxels \cite{Yu2021Plenoxels}, INGP \cite{Muller2022InstantNeuralGraphics}, Mip-NeRF360, and 3DGS.

\noindent\textbf{Results.} \cref{tab:5d_exp} presents a quantitative comparison between WIPES and leading 5D novel view synthesis approaches, demonstrating that our approach outperforms existing methods in most scenarios. Notably, WIPES exhibits significant advantages in environments with complex frequency components, such as large-scale outdoor scenes. Additionally, \cref{fig:5d_exp} provides qualitative comparisons for selected indoor and outdoor scenes from the dataset. The results indicate that WIPES excels at reconstructing reflective environments (\eg, mirror frames and truck windshields) and effectively captures high-frequency details (\eg, potted plants and distant transmission towers).

\noindent\textbf{Comparison with frequency-constrained approaches.} We compare WIPES with Gaussian-based approaches that utilize frequency priors or guidance. Since FreGS \cite{Zhang2024FreGS} has not released its source code, we present quantitative results against GES \cite{Hamdi2024GES}. GES employed a densify gradient threshold of $3e^{-4}$, whereas the original 3DGS \cite{Kerbl2023GaussianSplatting} used $2e^{-4}$. To ensure fairness, we evaluated GES with both thresholds. Experiments on the Mip-NeRF360, Tanks $\text{\&}$ Temples, and DeepBlending datasets (\cref{tab:5d_vs_ges}) demonstrate that WIPES achieves superior reconstruction without requiring frequency guidance. Additionally, when using identical densify gradient thresholds, WIPES matches GES in parameter count, highlighting its ability to deliver enhanced results without relying on frequency priors or guidance while maintaining comparable parameter efficiency.

\subsection{6D Dynamic Novel View Synthesis}

\noindent\textbf{Configurations.} In this section, we employed the D-NeRF \cite{Pumarola2020DNeRF} and NeRF-DS \cite{Yan2023NeRFDS} datasets to assess performance under varying dynamic conditions. WIPES utilized a frequency coefficient learning rate of $2.5 \times 10^{-3}$, consistent with the settings of D-3DGS \cite{Yang2024Deformable3DGaussians}. All other learnable parameters were maintained in alignment with D-3DGS to ensure a fair comparison. Differentiable rasterization was achieved through the integration of custom CUDA kernels, mirroring the configurations used in the 6D tasks.

\begin{figure*}[!htbp]
    \centering
    \vspace{-1em}
    \includegraphics[width=0.98\textwidth]{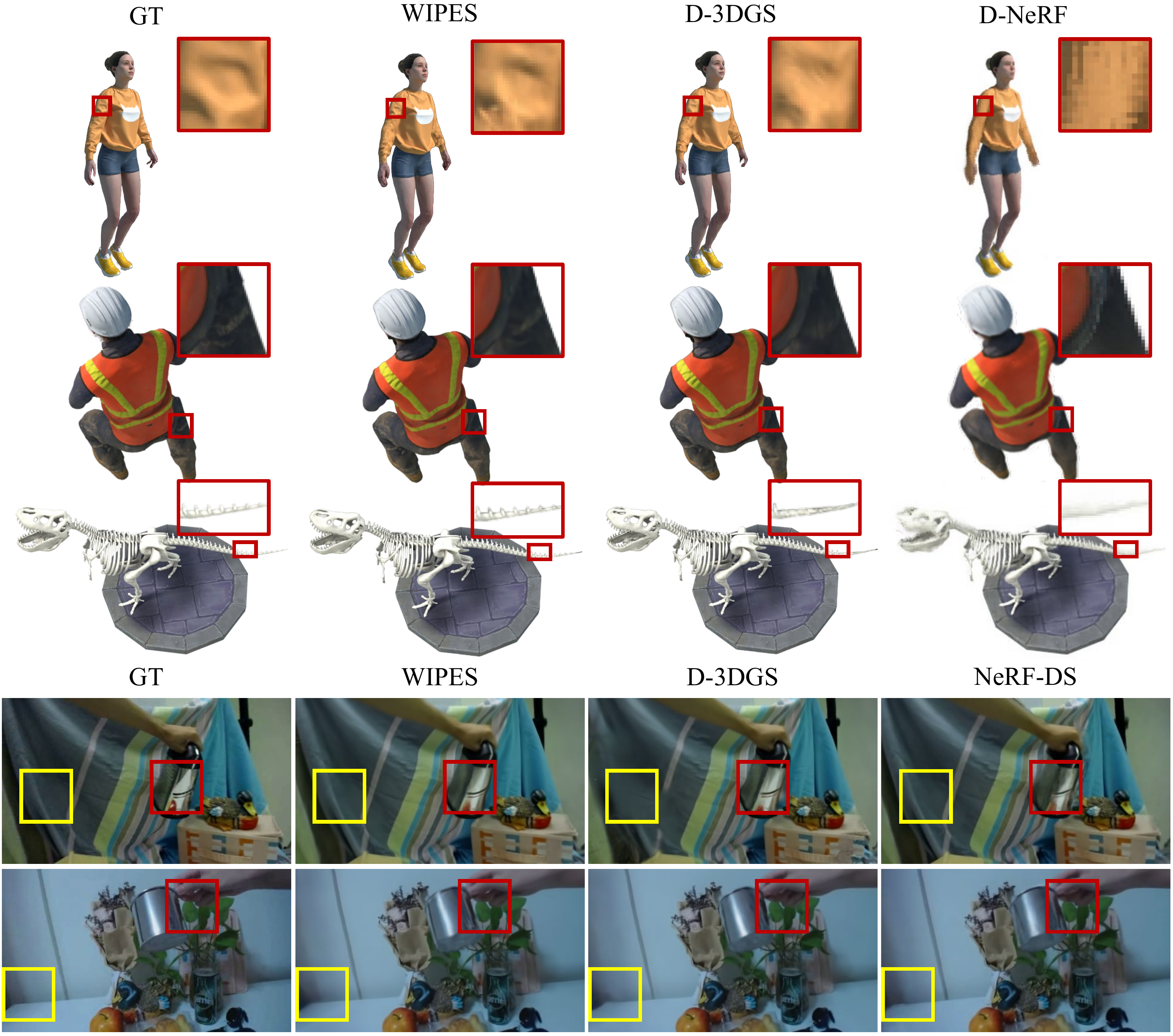}
    \vspace{-0.8em}
    \caption{Qualitative comparison on D-NeRF and NeRF-DS datasets. Our approach demonstrates more accurate registration, delivering outstanding performance in both static and dynamic regions.}
    \vspace{-1.6em}
    \label{fig:6d_exp}
\end{figure*}

\noindent\textbf{Results.} \cref{tab:6d_exp} presents the quantitative experimental results of WIPES in comparison with INR-based approaches and Gaussian-based approaches referenced in our work. WIPES exhibits enhanced expressive capabilities for various frequency components in dynamic scenes, enabling similar reconstruction performance to D-3DGS with fewer primitives. This indicates that the deformable INR in WIPES more effectively learns the dynamic features of primitives. Consequently, WIPES achieves superior performance in both reconstruction quality and rendering speed. \cref{fig:6d_exp} showcases qualitative results of WIPES across different dynamic datasets. It is evident that WIPES delivers more detailed and accurate results for both low-frequency elements, such as cloth, and high-frequency textures.

\begin{table*}[!htbp]
    \centering
    \setlength\tabcolsep{13.1pt}
    \caption{Quantitative comparison of dynamic novel view synthesis on D-NeRF and NeRF-DS datasets.}
    \vspace{-1em}
    \label{tab:6d_exp}
    \small
    \begin{tabularx}{\textwidth}{@{}p{2.5cm}|*{4}{c}|*{4}{c}@{}} 
    \toprule
    \multirow{2}{*}{Methods$\mid$Metrics} & \multicolumn{4}{c|}{\textbf{D-NeRF}} & \multicolumn{4}{c}{\textbf{NeRF-DS}} \\ 
     & PSNR$\uparrow$ & SSIM$\uparrow$ & LPIPS$\downarrow$ & FPS$\uparrow$ & PSNR$\uparrow$ & SSIM$\uparrow$ & LPIPS$\downarrow$ & FPS$\uparrow$ \\ 
    \midrule
    D-NeRF \cite{Pumarola2020DNeRF} & 30.43 & 0.9572 & 0.0724 & 0.117 & -- & -- & -- & -- \\
    NeRF-DS \cite{Yan2023NeRFDS} & -- & -- & -- & -- & 23.58 & \cellcolor{pink!95}0.8582 & 0.2496 & 0.008 \\
    D-3DGS \cite{Yang2024Deformable3DGaussians} & \cellcolor{pink!30}39.16 & \cellcolor{pink!30}0.9895 & \cellcolor{pink!30}0.0132 & \cellcolor{pink!30}71.7 & \cellcolor{pink!30}23.69 & 0.8423 & \cellcolor{pink!30}0.1976 & \cellcolor{pink!30}38.36 \\
    \textbf{WIPES} & \cellcolor{pink!95}39.52 & \cellcolor{pink!95}0.9899 & \cellcolor{pink!95}0.0127 & \cellcolor{pink!95}84.0 & \cellcolor{pink!95}23.95 & \cellcolor{pink!30}0.8527 & \cellcolor{pink!95}0.1762 & \cellcolor{pink!95}42.45 \\
    \bottomrule
    \end{tabularx}
\end{table*}

\section{Conclusion}
\label{sec:conclusion}
In this paper, we have proposed WIPES, a universal wavelet-based visual primitive for representing multi-dimensional visual signals. We have enhanced the Morlet function by designing a universally applicable and continuously differentiable wavelet primitive. Without relying on any frequency guidance or priors, WIPES has adaptively captured various real-world visual signals with rich frequency components, achieving impressive results in reconstruction tasks for 2D image representation, 5D static, and 6D dynamic novel view synthesis. To facilitate WIPES, we have developed a fast differentiable wavelet rasterizer that seamlessly integrates into Gaussian-based pipelines, enhancing rendering speed and efficiency during reconstruction. We have conducted extensive experiments on several challenging datasets across different dimensions, demonstrating the generality and effectiveness of our approach through both qualitative and quantitative evaluations.

However, our framework has limitations. The current Gaussian-based training pipeline has restricted scene representation due to fixed hyperparameters. Additionally, gradient-based densification in Gaussian approaches has caused instability and performance variability. Future work will aim to develop a more robust and efficient wavelet-based training and rendering framework to address these issues, fully leveraging WIPES's potential and advancing continuous signal representation in computer vision.
\small
\bibliographystyle{ieeenat_fullname}
\bibliography{main}


\end{document}